\documentclass[11pt,a4paper]{article}
\usepackage[hyperref]{ranlp2021}
\usepackage{times}
\usepackage{latexsym}

\usepackage{microtype}
\usepackage{algorithmic, algorithm}
\usepackage{enumitem}
\setlist{labelsep=5mm, leftmargin=5mm} % or \setlist{noitemsep} to leave space around whole list
\usepackage{graphicx}
\usepackage{soul}
\usepackage{multirow}
\usepackage{dblfloatfix}    % To enable figures at the bottom of page
\usepackage{amsmath}

\aclfinalcopy % Uncomment this line for the final submission
\title{Decoupled Transformer for Scalable Inference \\ in  Open-domain Question Answering}

\author{Haytham ElFadeel \\ \texttt{haythamf@fb.com} \\\And Stan Peshterliev \\ \texttt{stanvp@fb.com}
    \\}

\date{}

\begin{document}
\maketitle
\begin{abstract}
Large transformer models, such as BERT, achieve state-of-the-art results in machine reading comprehension (MRC) for open-domain question answering (QA). However, transformers have a high computational cost for inference which makes them hard to apply to online QA systems for applications like voice assistants. To reduce computational cost and latency, we propose decoupling the transformer MRC model into input-component and cross-component. The decoupling allows for part of the representation computation to be performed offline and cached for online use. To retain the decoupled transformer accuracy, we devised a knowledge distillation objective from a standard transformer model. Moreover, we introduce learned representation compression layers which help reduce by four times the storage requirement for the cache. In experiments on the SQUAD 2.0 dataset, a decoupled transformer reduces the computational cost and latency of open-domain MRC by 30-40\% with only 1.2 points worse F1-score compared to a standard transformer.
\end{abstract}

\section{Introduction}

Open-domain question answering (QA) aims to answer questions from a collection of text passages. It is an important and challenging task with application to several domains such as web search and voice assistants. The most popular architecture for open-domain QA is \textit{retriever}-\textit{reader}~\citep{chen2017reading}. Given a question, the retriever uses an information retrieval (IR) system over a collection of passages to return top-K results that are most likely to contain an answer. Then, the reader uses a machine reading comprehension (MRC) model on each of the top-K results to find an answer. In the end, the top-K MRC answers are ranked to produce a final answer. 

% Since BERT was introduced, transformers have revolutionized the natural language processing with many improvements to MRC, natural language inference (NLI), and more broadly a resounding impact on the way contemporary language modeling is carried out

For both the retriever and the reader, large transformer models such as BERT~\citep{devlin2018bert}, RoBERTa~\citep{liu2019roberta} and ELECTRA~\citep{clark2020electra} achieve state-of-the-art results. A disadvantage of large transformer models is the high computational cost for inference which makes them hard to apply to online runtime systems, e.g. voice-assistants. Transformers' computational cost comes from three major factors. Firstly, the size of the feed-forward layers which project to an intermediate higher dimension and projects back to the original dimension. Secondly, the multi-head self-attention has quadratic computational complexity in the sequence length. Thirdly, the total number of layers.

Dense passage retrieval (DPR)~\citep{karpukhin2020dense} is a retriever model which uses a transformer question encoder and transformer passage encoder to capture semantic similarity. The question and passage encoders are trained such that passages that are likely to contain an answer have a large embedding dot product with the question embedding. The embeddings for the passages are generated offline and indexed for efficient distributed KNN search~\citep{JDH17}, and only the embedding for the question is generated at runtime. Since questions are usually short, the retriever runtime inference computational cost is low.

MRC reader models process the top-K passages returned by the retriever to get an answer. In transformer-based MRC models, each passage is encoded together with the question using a CLS and separator characters \texttt{[CLS] Question [SEP] Passage}. The encoding is followed by a prediction head which determines the answer span. If there is no answer, the model result is a zero-length span on the CLS token. The joint encoding of the document and the question produces rich interaction features but it increases the sequence length, and thus the computational coast. Since the MRC model inference is executed at runtime on long sequences of a question and passages, MRC is the main computational bottleneck in retriever-reader QA.

\begin{figure*}[!th]
    \centering
    \includegraphics[page=1,clip, trim=2cm 1cm 4cm 7cm, width=1.00\linewidth]{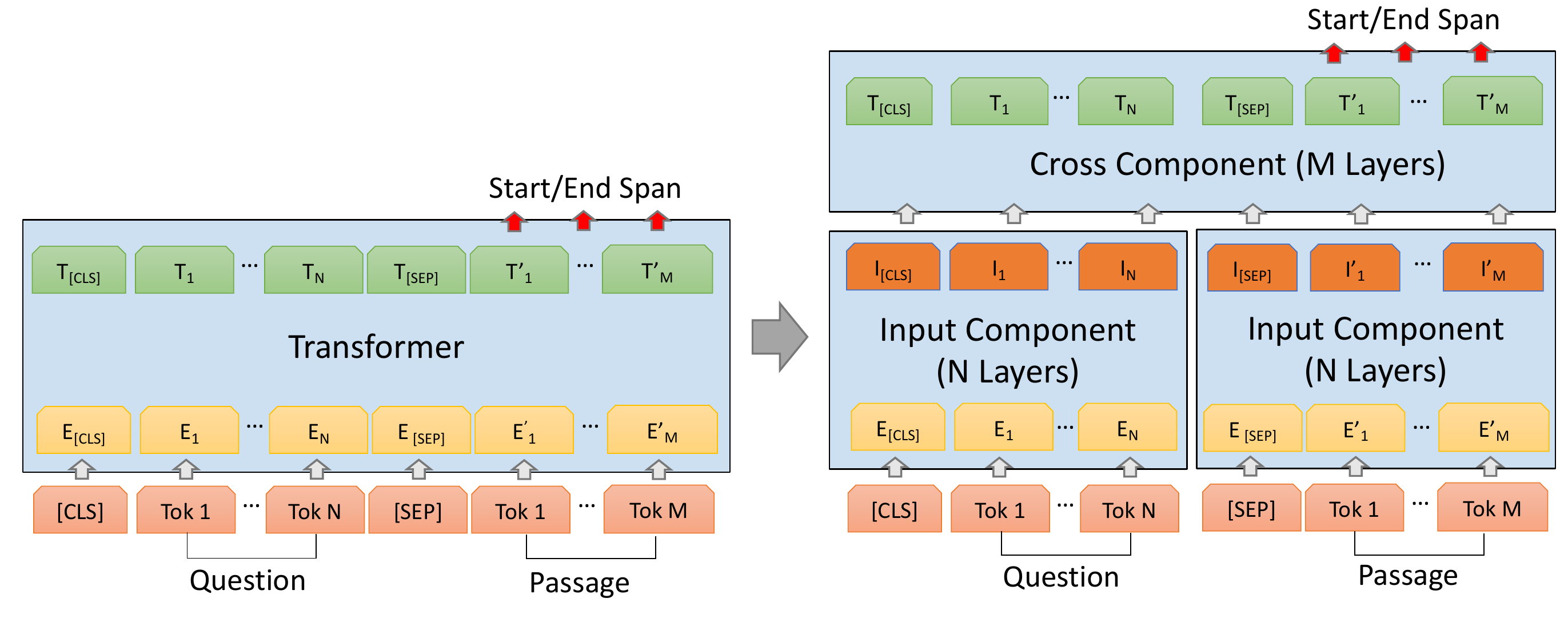}
    \caption{On the left, standard transformer model for MRC. On the right, decoupled transformer model with input-component and cross-component.}
    \label{fig:decoupled_model}
    \vspace{-3mm}
\end{figure*}

There have been several ideas to reduce the runtime inference of transformer models, such as precision reduction via quantization~\citep{zafrir2019q8bert, shen2020q}, knowledge distilling to a smaller architecture~\citep{sanh2019distilbert, jiao2019tinybert}, and approximate multi-head attention for reducing the quadratic complexity~\citep{wang2020linformer, beltagy2020longformer}. In this paper, we take an orthogonal approach, decoupling the transformer encoding for multiple inputs to improve efficiency, which can be combined with the aforementioned techniques. The motivation for the decoupled transformer is that in open-domain QA the passages are known in advance and part of the passage computation can be performed offline and stored. Then, online at the runtime the question computation can be performed only once and combined with the stored state from passages with cross-attention.

We use the decoupled transformer to reduce the computational cost of open-domain MRC by 30-40\% with only 1.2 points worse than the F1-score on the SQUAD 2.0 benchmark.

Our contributions are as follows:

\begin{itemize}
    \item We propose and evaluate a novel decoupled transformer approach for MRC in open-domain QA to reduce runtime inference cost. Our approach uses a knowledge distillation (KD) objective to bridge the gap between a standard transformer and decoupled transformer. 
    \item We conduct experiments to understand how much cross-attention between inputs is needed in MRC and other natural language processing (NLP) tasks like paraphrasing identification and natural language inference. 
    \item We devise an accurate representation compression approach to reduce the storage requirement for decoupled transformer offline state. The compression provides a four-fold reduction in the index storage requirement for large corpora such as Wikipedia, from 3.4 TB to 858 GB.
\end{itemize}

\section{Related Work}

DC-BERT~\citep{zhang2020dc} is a decoupled transformer that has dual BERT models. An online BERT encodes the question and an offline BERT pre-encodes all the passages and stores their encodings. Conceptually, DC-BERT goals and architecture of combining local and global context are similar to our work with the following major differences:
\begin{itemize}
    \item We apply decoupled transformers to MRC and DC-BERT is designed and evaluated for the retrieval passage ranking. With the recent introduction of DPR, passage ranking as explored in DC-BERT is less important, so MRC becomes the primary bottleneck.
    \item We investigate how much cross-attention is needed for MRC and other NLP tasks.
    \item We introduce compression and decompression layers to reduce representation storage requirements.
\end{itemize}

Another model where the query and the passage are encoded independently using a transformer is ColBERT~\citep{khattab2020colbert}. The main modeling applications for ColBERT are retrieval and passage ranking. After encoding the query and the passage independently, late interactions are introduced using an efficient sum of maximum similarity computations. ColBERT representations are used for retrieval, so it combines the strengths of DPR and DC-BERT. However, the efficient late interactions in ColBERT do not have enough representation power for complex tasks like MRC.

\section{Decoupled Transformer}

% In tasks where part of the transformer inputs doesn’t change often or could be cached, such as: Document Ranking in Information Retrieval, where the documents representations doesn’t change often and could be cached. Question Answering - Machine Reading Comprehension, where we can compute and reuse the passages representations, Natural Language Inference and Similarity matching. The decoupled transformer aims to reduce the inference computation by processing the inputs independently for part of the process and eliminating redundancy computation.

In the decoupled transformer, Figure~\ref{fig:decoupled_model}, we split the transformer model $M$ into two components.
\begin{enumerate}
    \item Input-component $M_{input}$ (the lower $N$ layers) which processes the inputs independently and produces a representation. The representation for the inputs that are known in advance, i.e. the passages, is stored and used without computation.
    \item Cross-component $M_{cross}$ (the higher M layers) which processes the inputs jointly (after concatenation) and produces the final output.
\end{enumerate}

\subsection{Workflow}

\begin{figure}[!ht]
    \centering
    \includegraphics[page=2,clip, trim=16cm 0.7cm 4cm 1cm, width=1.00\linewidth]{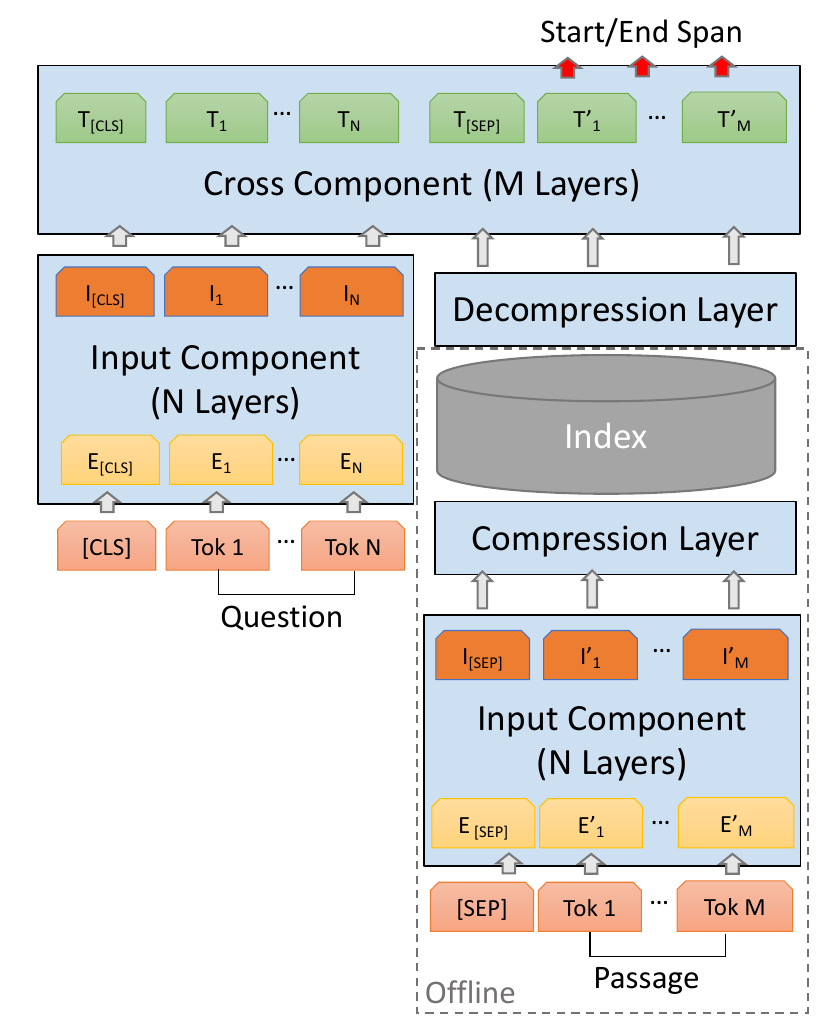}
    \caption{Decoupled model offline indexing and runtime. The passage is processed with the input module, then the state representations are compressed using a projection layer and stored in the index. At runtime, the passage is retrieved by DPR from the index, decompressed to the original representation size and used by the cross-component. }
    \label{fig:decoupled_model_runtime}
    \vspace{-3mm}
\end{figure}

The workflow is depicted in Figure~\ref{fig:decoupled_model_runtime}. Offline, we run the input-component $M_{input}$ on each passage from the collection of passages and store the representation in the search index. Moreover, we compress the stored passage representation to reduce storage requirements. The offline step is performed together with the DPR indexing.

At runtime, using DPR we retrieve the candidate passages with their stored representation which we decompress. Then, we compute the question representation using the input-component $M_{input}$. Finally, we concatenate the question representation with the representation of the passage and process them with the cross-component $M_{cross}$.

\subsection{Benefits}

The decoupled transformer reduces per question transformer complexity in lower $N$ layers from $O(N_p (L_q + L_p)^2)$ to $O(L^2_q + N_p L^2p)$ where $N_p$ denotes the number of top-K passages per question, $L_q$ and $L_p$ denote the average number of tokens of each question and passage.

At runtime, the computation for the lower $N$ layers for the passage is eliminated because it is performed once offline and reused. Moreover, the computation for the lower $N$ layers for the question is done only once for the top-K retrieved passages, and not repeated, as opposed to the normal transformer which uses all layers on both the question and the top-K retrieved passages. 

\subsection{Initialization}

To build a decoupled transformer model, we start from a standard transformer such as BERT, RoBERTa, ELECTRA model which is fine-tuned on a target dataset such as SQUAD 2.0. Then, we create the decoupled transformer model by splitting the encoder layers into input and cross components which are initialized with the fine-tuned MRC model weights. In addition to the standard transformer weights, we create a global position embedding and segment embedding layers at the start of the cross-component and initialize them to the same weight as the local position and segment embedding from the input-component. The global position and segment embedding re-encode the tokens for the new position in the concatenated question-document encoding sequence. The segment embedding differentiates whether the encoded token is from the question or document.

\subsection{Training Objective}

During decoupled transformer training, we aim to preserve the standard transformer model accuracy. To achieve that, we propose a knowledge distillation (KD)~\citep{hinton2015distilling} objective from the standard transformer to decoupled transformer which helps preserve the original representation.

The objective function is the sum of four terms: 
\begin{align}
L &= (1 - \lambda)CE(\text{y}, \text{target}) \\
  &+ \lambda KL(\text{logits} / \text{T}, \text{teacher-logits} / \text{T}) \\
  &+ \sigma MSE(\text{represent}_n, \text{teacher-represent}_n) \\
  &+ \sigma MSE(\text{attention}_n, \text{teacher-attention}_n)
\end{align} 

\begin{enumerate}
    \item A standard cross-entropy (CE) loss with the prediction $y$ and hard targets from ground truth labels.
    
    \item KD loss based on Kullback–Leibler (KL) divergence with logits from the teacher standard transformer model. We scale the targets with the same temperature $T$ for both the teacher and student.
    
    \item The mean square error (MSE) between the decoupled model final layer representation with the original model final layer representation. 
    
    \item The MSE between the decoupled model final layer multi-head self-attention output with the standard model final layer multi-head self-attention output.
\end{enumerate}

The parameter $\lambda$ determines the relative contribution of CE and KL losses. And, $\sigma$ is a weight for the MSE losses. 

The MSE losses on the final layer representation and the final layer self-attention are similar to TinyBERT~\citep{jiao2019tinybert} approach to smaller model distillation. Unlike TinyBERT, we only apply the MSE losses only on the last layer. The motivation for the MSE losses is that we are aiming to make the representation at the end of the decoupled transformer to match the representation of the standard transformer.

%\hl{Haytham: As shown in the ablation study table, KD logits get us so far. feature based KD is not very useful (tricky/hard to improve upon regular logits based KD), in order for feature KD to be useful I applied it only to the last layer and needed to scale the loss down so that the model pay more attention to the loss signal from the logtis KD compared to the signal from feature KD, but using 0.5 for feature KD make all losses balanced, This is hyperparameter and I tried several option from 1.0 / 1.0, 0.5 / 0.5, 0.75 / 0.25, 0.25 / 0.75, etc.}  )

%We use Adam optimizer and the hyperparameters are listed in appendix 1. 
%\hl{Stan: we should put the hyperparameters here. We have 8 pages of content}
%\hl{haytham: In the google doc}

\section{Representation Compression}

In open-domain QA the collection of passages are known in advance. So, with decoupled transformer, we run the input-component $M_{input}$ on each passage offline and store the passage representation in the index. For a large corpus, the representation storage can be a significant amount. In the case of QA over Wikipedia, the storage requirement is around 3.4TB given around 32 million passages, averaging 150 tokens per passage, and 768 token dimensions for the BERT-base model with 16-bit precision.

To reduce the storage requirements for the passage representation of the decoupled transformer, we introduce a \textit{compression layer} at the end of the input-component and a \textit{decompression layer} at the start of the cross-component, see Figure~\ref{fig:decoupled_model_runtime}. The compression layer is a linear projection from the original dimension to a compression dimension. The decompression layer is a linear projection from the compression dimension to the original dimension. These layers are similar to an autoencoder with a bottleneck. 

\subsection{Training Procedure}

To train the compression and decompression layers we start from a decoupled transformer model. Then, we perform training in two phases:

\begin{itemize}
\item \textbf{Phase 1.} We train the randomly initialized compression and decompression layers to reconstruct the input-component output representation without updating the decoupled transformer model itself. 
\item \textbf{Phase 2.} We train the compression and decompression layers together with the decoupled transformer jointly. This means the cross-component receives the decompressed representation.
\end{itemize}

The intuition behind the two-phase approach is that since compression and decompression layers are randomly initialized, it is beneficial to first train the compression and decompression layers independent from the decoupled transformer to get near-optimal weights. Then, train the cross-component of the model to understand the slightly different decompressed representation.

\section{Experiments and Results}

\begin{table*}[!b]
\begin{tabular}{lcccccccccccc}
\hline
\textbf{Task} & \textbf{Baseline} & \textbf{1-11} & \textbf{2-10} & 
\textbf{3-9} & \textbf{4-8} & \textbf{5-7} & \textbf{6-6} & \textbf{7-5} & \textbf{8-4} & \textbf{9-3} & \textbf{10-2} & \textbf{11-1} \\
\hline
SQUAD 2.0 & 87.6 & 87.5 & 87.2 & 87.1 & 87.0 & 86.7 & 85.4 & 85.2 & 84.8 & 84.0 & 80.6 & 62.6 \\
QQP & 91.5 & 91.1 & 91.0 & 90.9 & 90.9 & 90.9 & 90.5 & 90.4 & 90.4 & 90.0 & 89.0 & 86.4 \\
MNLI & 88.9 & 87.3 & 87.1 & 86.9 & 86.7 & 86.7 & 86.7 & 86.4 & 86.4 & 85.6 & 77.0 & 73.5 \\
MRPC & 89.5 & 87.7 & 86.3 & 85.5 & 83.0 & 80.1 & 78.1 & 77.9 & 77.8 & 77.7 & 71.8 & 71.6 \\
FLOPs & 1.0 & .91 & .83 & .75 & .66 & .58 & .50 & .41 & .33 & .25 & .16 & 0.08 \\
\hline
\end{tabular}
\caption{Decoupled transformer results with variable number of input-component and cross-component layers. Baseline is a standard transformer  model. The $x$-$y$ columns indicate the number of input-component and cross-component layers. We use F1 score for SQUAD 2.0 and accuracy for QQP, MNLI, and MRPC. There is a consistent trend for performance to degrade as we increase the number of input-component layers and decrease the number of cross-component layers. FLOPs is floating point operations for inference as a measure of computational cost.}
\label{tbl:layer_results}
\end{table*}

\subsection{Datasets}

We evaluate the decoupled transformer on SQUAD 2.0~\citep{rajpurkar2018know} which is a popular MRC dataset over Wikipedia articles. In addition to MRC, we evaluate models on the datasets below to understand how many cross-components layers are needed for tasks of different complexity and dataset size.

\begin{itemize}
    \item QQP~\citep{chen2018quora} and MRPC~\citep{dolan2005automatically} datasets for paraphrasing identification. The task is given two sentences, recognizing if they are paraphrases or not.
    \item MNLI~\citep{williams2018multi} dataset for natural language inference datasets. The task is given two sentences the ``premise'' and the ``hypothesis'', to determine if the hypothesis entails, contradicts, or is neutral given the premise.
\end{itemize}

\begin{table}[!ht]
\begin{tabular}{lc}
\hline
\textbf{Hyperparameter} & \textbf{Value} \\ \hline
Warmup steps & 200 \\
Learning Rate (LR) & 5e-5 \\
Layer-wise LR Multiplier & 0.95 \\
Batch size per GPU & 32 \\
Number of GPUs & 2 \\
Adam $\epsilon$, $\beta_1$, $\beta_2$ & 1e-6, 0.9, 0.999 \\
Attention Dropout & 0.1 \\
Dropout & 0.1 \\
Weight Decay & Linear \\
Gradient Clipping & 3.0 \\
Epochs & 4 \\
KD temperature $T$ & 3.0 \\
Loss weight $\lambda$ & 0.95 \\
Loss weight $\sigma$ & 0.5 \\ \hline
\end{tabular}
\caption{Model training hyperparameters.}
\label{tbl:hyperparamters}
\end{table}

\subsection{Setup}

\textbf{Models.} We use ROaD-base~\citep{elfadeel2021robustly} for the MRC experiments on SQUAD 2.0. ROaD is an ELECTRA model pretrained and distilled using multi-task learning. For the experiments on QQP, MRPC, and MNLI we use ELECTRA-base. All models are implemented in PyTorch and optimized using Adam~\citep{kingma2014adam}.

\textbf{Hyperparameters.} Table~\ref{tbl:hyperparamters} shows the hyperparameters that we use for fine-tuning the standard transformer and training the decoupled transformer. We searched different values for the temperature $T$, and the weights $\lambda$ and $\sigma$. For $\lambda$, we experimented with 0.5, 0.7, 0.9, 0.95 values and we found that for decoupled transformer training a large $\lambda$ that biases towards the KL divergence objective work best. For $\sigma$, we experimented with $0.25, 0.5, 0.75, 1.0$ values and we found that smaller values work better because otherwise the KL divergence objective is given less weight which leads to worse models.

\textbf{Hardware.} We perform the experiments and benchmarks on Nvidia Titan RTX with tensor cores GPU and AMD Ryzen Threadripper 3960X - 24 cores CPU.

%\hl{Haytham: As shown in the ablation study table, KD logits get us so far. feature based KD is not very useful (tricky/hard to improve upon regular logits based KD), in order for feature KD to be useful I applied it only to the last layer and needed to scale the loss down so that the model pay more attention to the loss signal from the logtis KD compared to the signal from feature KD, but using 0.5 for feature KD make all losses balanced, This is hyperparameter and I tried several option from 1.0 / 1.0, 0.5 / 0.5, 0.75 / 0.25, 0.25 / 0.75, etc.}  )

\subsection{Decoupled Transformer}

%\hl{Have to better motivate why use non-QA datasets given that they don't really have static inputs. Short description of each dataset/task is going to be useful. Did we try something that didn't work so well?}
%\hl{Haytham: The real motivation is to understand how much interaction (cross-attention) is really needed between the inputs. Humans we can comprehend information then asked questions about them, so why can't we do something similar. as in the results it seems all datasets follow similar trajectory but MRPC degrade quickly - maybe because the dataset is smaller.}

First, we perform a set of experiments on a decoupled transformer without compression. For each experiment, we denote the decoupled transformer split configuration as $x$-$y$, where $x$ is the number of input-component layers, and $y$ is the number of cross-component layers.

Table~\ref{tbl:layer_results} shows the performance and FLOPs starting from the baseline standard transformer model to decoupled transformer with extreme 11-1 split. We observe that tasks with a large dataset (QQP, SQUAD, MNLI contain over 100K samples each) have similar behavior with a noticeable drop when moving from decoupled transformer 5-7 split to 6-6 split, and another big drop when the number of cross-component layers becomes less than 3. While in MRPC, a small dataset with around 5K sample, the drop of performance was significant and bigger than the other large datasets even with the decoupled transformer with 1-11 split.

With every layer, we moved from the cross-component to the input-component, the FLOPs decreased by about 8\% and performance dropped by a small amount until the number of input-component layers equals to or bigger than the cross-component layers. The results show that choosing the right setting is application-specific and the best option depends on the particular performance and latency trade-offs. 

For the following experiments, we use the 5-7 split because it provides the best trade-offs between accuracy and FLOPs across the evaluated datasets.

\begin{table}[!ht]
{\renewcommand{\arraystretch}{1.1}% for the vertical padding
\begin{tabular}{lll}
\hline
\multirow{2}{*}{\textbf{MRC Model}} & \multicolumn{2}{c}{\textbf{SQUAD 2.0}} \\
 & \multicolumn{1}{c}{\textbf{F1}} & \multicolumn{1}{c}{\textbf{EM}} \\ \hline
Decoupled 5-7 & 86.7 & 84.1 \\
- SQUAD 2.0 pretraining & 84.2 & 81.5 \\
\begin{tabular}[c]{@{}l@{}}- training position and segment \\ \hspace{3px} embedding in the cross-model\end{tabular} & 82.1 & 80.0 \\
- KL objective & 84.0 & 80.4 \\
\begin{tabular}[c]{@{}l@{}}- MSE on representation and \\ \hspace{3px} attention final layer\end{tabular} & 86.5 & 83.7 \\
\begin{tabular}[c]{@{}l@{}}+ MSE on hidden and \\ \hspace{3px} attention applied to all layers\end{tabular} & 86.0 & 83.2 \\
\hline
\end{tabular}
}
\caption{Decoupled transformer ablation study for SQUAD 2.0 MRC decoupled transformer with 5-7 split. We remove one row at a time except for the last row where we add MSE losses to all layers and not just the final layer.}
\label{tbl:decopuling_ablation}
\end{table}

\textbf{Ablations}. We perform an ablation study to understand the effect of the different modeling techniques on the decoupled transformer performance. Table~\ref{tbl:decopuling_ablation} shows the results. First, we remove the SQUAD 2.0 pretraining and start with regular ELECTRA-base which reduces F1 significantly by 2.5 points. Then, we tried keeping the position and segment embeddings in the cross-component frozen which hurt F1 as expected. If we remove the distillation KL objective, F1 degrades significantly by 2.7 points. On the other hand, removing the MSE losses on the representation and attention does not cause a significant reduction in F1. However, adding MSE losses on all layers actually causes a reduction in F1 because the CE and KL objectives receive less weight.

\begin{table}[!ht]
\begin{tabular}{lcll}
\hline
\multirow{2}{*}{\textbf{Compression Rate}} & \multirow{2}{*}{\textbf{Dim Size}} & \multicolumn{2}{l}{\textbf{SQUAD 2.0}} \\
 &  & \textbf{F1} & \textbf{EM} \\
 \hline
No compression & 768 & 86.7 & 84.1 \\
2.0x & 386 & 86.6 & 84.0 \\
3.0x & 256 & 86.5 & 83.8 \\
4.0x & 192 & 86.4 & 83.8 \\
4.8x & 160 & 86.2 & 83.6 \\
6.0x & 128 & 85.2 & 82.4 \\
\hline
\end{tabular}
\caption{Decoupled transformer compression rates for SQUAD 2.0 MRC decoupled transformer with 5-7 split. The dim size is the number of embedding dimensions per token. EM is exact match accuracy. FLOPs is floating-point operations for inference as a measure of computational cost.}
\label{tbl:compression_rate}
\end{table}

\subsection{Compression}

To evaluate compression, we conducted experiments on a decoupled transformer with 5-7 split using the MRC model for SQUAD 2.0. Our goal is to understand how much impact different levels of compression have on the storage requirement and model performance.

Table~\ref{tbl:compression_rate} compares the results with five different levels of compression. We observed the performance degradation is minimal for 2x, 3x and 4x compression, and then it starts to degrade significantly. At 4x compression, the required storage for open-domain QA over Wikipedia with the previous assumptions is 3.4 TB which could be reduced to 858 GB.

\textbf{Ablations}. We evaluate the effectiveness of the two-stage training of the compression and decompression layers. Table~\ref{tbl:compression_ablation} shows the results. First, we remove the training of the compression independent from the model fine-tuning which causes a significant 6.6 F1 score reduction. Second, we remove the joint training of compression and MRC layers which cause 1.6 F1 sore drops. Overall, both stages are necessary for training effective compression layers.

\begin{table}[!ht]
{\renewcommand{\arraystretch}{1.1}% for the vertical padding
\begin{tabular}{lll}
\hline
\multirow{2}{*}{\textbf{MRC Model}} & \multicolumn{2}{c}{\textbf{SQUAD 2.0}} \\
 & \multicolumn{1}{c}{\textbf{F1}} & \multicolumn{1}{c}{\textbf{EM}} \\ \hline
Decoupled 5-7, 4x compress & 86.4 & 83.8 \\
\begin{tabular}[c]{@{}l@{}}- training compression \\ \hspace{3px} independent from the model\end{tabular} & 79.8 & 77.3 \\
- joint training & 84.8 & 82.3 \\ \hline
\end{tabular}
}
\caption{Compression ablation study for MRC SQUAD 2.0 model with 5-7 split. We remove either training compression independent of the model or join training.}
\label{tbl:compression_ablation}
\end{table}

\subsection{Inference Performance}

In addition to FLOPs computational cost analysis, we run inference benchmarks on GPU and CPU. For the benchmarks, we use FP16 PyTorch models without TorchScript. We test in two settings: long and short inputs. Long inputs are 64 words for the question and 448 words for the passage. Short inputs are 16 words for the question and 150 words for the passage. For each setting, we perform four runs and take the average time.

Table~\ref{tbl:real_latency} shows the benchmark results. For CPU the results are close to our FLOPs analysis, and for GPU the we get lower runtime reduction due to the GPU parallelism.

\begin{table}[!ht]
\centering
\small
\setlength{\tabcolsep}{0.3em} % for the horizontal padding
{\renewcommand{\arraystretch}{1.3}% for the vertical padding
\begin{tabular}{lcc}
\hline
\multirow{2}{*}{\textbf{MRC Model}} & \textbf{GPU} & \textbf{CPU} \\
 & \textbf{Long / Short (diff)} & \textbf{Long / Short (diff)} \\
\hline
Baseline & 9.1 / 9.0 & 3200 / 920 \\
Decoupled 5-7 & 6.3 / 6.2 (31\%) & 1890 / 490 (40-46\%) \\
\hspace{2px}+ 4x compress & 6.4 / 6.3 (30\%) & 1950 / 520 (39-43\%) \\ 
\hline
\end{tabular}
}
\caption{Decoupled transformer inference performance. Baseline is a standard transformer MRC model. Long and short indicate the input length, and the times are in milliseconds. Diff is the difference with the baseline. }
\label{tbl:real_latency}
\end{table}

\begin{table}[!ht]
\begin{tabular}{lcll}
\hline
\multirow{2}{*}{\textbf{Model}} & \multirow{2}{*}{\textbf{FLOPs}} & \multicolumn{2}{l}{\textbf{SQUAD 2.0}} \\
 &  & \textbf{F1} & \textbf{EM} \\ \hline
DeBERTa & 1.2x & 86.2 & 83.1 \\
ROaD & 1.0x & 87.6 & 85.1 \\
\begin{tabular}[c]{@{}l@{}}Decoupled ROaD\\ 5-7, 4x compress\end{tabular} & 0.6x & 86.4 & 83.8 \\
\hline
\end{tabular}
\caption{Decoupled transformer with DeBERTA-base and ROaD-base on SQUAD 2.0 model. EM is exact match accuracy. FLOPs is floating point operations for inference as a measure of computational cost.}
\label{tbl:deberta_comparison}
\end{table}

\subsection{Results}

Table~\ref{tbl:deberta_comparison} compares the decoupled ROaD-base 5-7 split model with 4x compression with the ROaD-base model and DeBERTa-base model on SQUAD 2.0 MRC task. The DeBERTa model introduces additional positional embeddings that increase the computational cost by 20\%. Still, the decoupled ROaD model achieves comparable accuracy with DeBERTa while requiring two times fewer FLOPs.

\section{Conclusion and Future Work}

We presented the decoupled transformer model for reducing runtime latency of MRC models in open-domain QA. The decoupling allows for part of the representation computation to be performed offline and cached for online use. To bridge the accuracy gap between a standard transformer and decoupled transformer, we devised knowledge distillation objectives for both model logits and features. Moreover, we introduced a representation compression approach that allows for a four-times reduction in representation storage requirements for open-domain QA without significant loss of accuracy. We use the decoupled transformer to reduce the computational cost of open-domain MRC by 30-40\% with only 1.2 points worse than the F1-score on the SQUAD 2.0 benchmark.

In the future, we are planning to extend the decoupled model with a DPR objective. The goal is for the input-component to also produce DPR-like embeddings suitable for similarity search. This way, we can have a single model that acts as both retrieval and reader.

\bibliographystyle{acl_natbib}
\bibliography{ranlp2021}

%\appendix

\end{document}